\documentclass[letterpaper, 10 pt, conference]{ieeeconf}  

\IEEEoverridecommandlockouts                              

\let\labelindent\relax
\usepackage{enumitem}
\usepackage{booktabs}
\usepackage{caption}
\usepackage{xcolor}
\usepackage{hyperref}
\hypersetup{colorlinks=true}
\usepackage{mathtools}
\usepackage{amsfonts}
\usepackage{mathtools}
\usepackage{lipsum}
\usepackage{multicol}
\usepackage{multirow}
\usepackage{subcaption}
\usepackage{tcolorbox}

\usepackage{tabularx}
\usepackage{colortbl}

\usepackage[subtle,title=tight]{savetrees}

\definecolor{dgreen}{rgb}{0,0,0}
\definecolor{dyellow}{rgb}{.7,.7,0}
\definecolor{dred}{rgb}{1,0,0}
\definecolor{dblue}{rgb}{0,0,0.7}
\definecolor{dorange}{rgb}{0.9,0.5,0.1}
\definecolor{light-gray}{rgb}{0.8, 0.8, 0.8}
\definecolor{highlight}{HTML}{e3eeff}
\definecolor{comment-green}{rgb}{0.435, 0.576, 0.106}
\definecolor{prompt-gray}{HTML}{a7a7a7}
\definecolor{code-syntax}{HTML}{0060b1}

\newcommand{\ie}{i.e., }
\newcommand{\eg}{e.g., }

\newcommand{\command}[1]{\textcolor{comment-green}{#1}}
\newcommand{\prompt}[1]{\textcolor{prompt-gray}{#1}}

\usepackage{inconsolata}
\usepackage[T1]{fontenc}

\newcommand{\hlcode}[1]{\colorbox{highlight}{\makebox[0.96\linewidth][l]{#1}}}

\newcommand{\lmp}[1]{
\begin{tcolorbox}[boxsep=0pt,
                  left=3pt,
                  right=-4pt,
                  top=3pt,
                  bottom=3pt,
                  arc=0pt,
                  boxrule=0.5pt,
                  colframe=light-gray,
                  colback=white
                  ]
\small{  
\ttfamily
#1
}
\end{tcolorbox}
}

\newcommand{\speciallmp}[1]{
\begin{tcolorbox}[
 enlarge top by=0.5em,
 boxsep=0pt,
                  left=3pt,
                  right=-4pt,
                  top=3pt,
                  bottom=3pt,
                  arc=0pt,
                  boxrule=0.5pt,
                  colframe=light-gray,
                  colback=white
                  ]
\small{  
\ttfamily
#1
}
\end{tcolorbox}
}

\title{\LARGE \bf
\ \ \\
\ \ \\
Audio Visual Language Maps for Robot Navigation
\vspace{0.75em}
}
\IEEEaftertitletext{\vspace{-1.25em}}
\author{Chenguang Huang$^{1}$, Oier Mees$^{1}$, Andy Zeng$^{2}$, Wolfram Burgard$^{3}$
\thanks{$^{1}$University of Freiburg, Germany.}
\thanks{$^{2}$Google Research, USA.}
\thanks{$^{3}$University of Technology Nuremberg, Germany.}
\vspace{-3em}
}
\date{}
\begin{document}

\maketitle
\thispagestyle{empty}
\pagestyle{empty}

\begin{abstract}
While interacting in the world is a multi-sensory experience, many robots continue to predominantly rely on visual perception to map and navigate in their environments. In this work, we propose Audio-Visual-Language Maps (AVLMaps), a unified 3D spatial map representation for storing cross-modal information from audio, visual, and language cues. AVLMaps integrate the open-vocabulary capabilities of multimodal foundation models pre-trained on Internet-scale data by fusing their features into a centralized 3D voxel grid. In the context of navigation, we show that AVLMaps enable robot systems to index goals in the map based on multimodal queries, e.g., textual descriptions, images, or audio snippets of landmarks. 
In particular, the addition of audio information enables robots to more reliably disambiguate goal locations.
Extensive experiments in simulation show that AVLMaps enable zero-shot multimodal goal navigation from multimodal prompts and provide 50\% better recall in ambiguous scenarios. These capabilities extend to mobile robots in the real world – navigating to landmarks referring to visual, audio, and spatial concepts. Videos and code are available at \href{https://avlmaps.github.io}{https://avlmaps.github.io}.
\end{abstract}

\section{INTRODUCTION}
Humans exhibit a remarkable ability to integrate and leverage multiple sensing modalities to efficiently move around in the physical world. Our actions are driven by a myriad of sensory cues: the sound of glass breaking might signal a dangerous situation, the microwave might buzz to indicate it is done, or a dog might bark to draw our attention. Acoustic signals particularly represent a valuable complementary form of information, also evident by the utility that it provides for the visually impaired who may rely on it for navigation. Research in cognitive science also suggests that children understand and integrate information from different sensing modalities into spatial cognitive maps \cite{kording2007causal}. In contrast, mobile robots today rely predominantly on visual perception to operate in human-centered environments. While recent advances in training modality-specific foundations models have spurred map representations that integrate pre-trained visual-language features for improved scene understanding~\cite{huang23vlmaps,chen2022open, shafiullah2022clipfields, jatavallabhula2023conceptfusion}, how to best spatially anchor additional sensing modalities, such as audio signals, in ways that enable effective data-efficient cross-modal reasoning for downstream robotics tasks, remains a relatively open question.
 \begin{figure}[t]
	\centering
	\includegraphics[width=1\columnwidth]{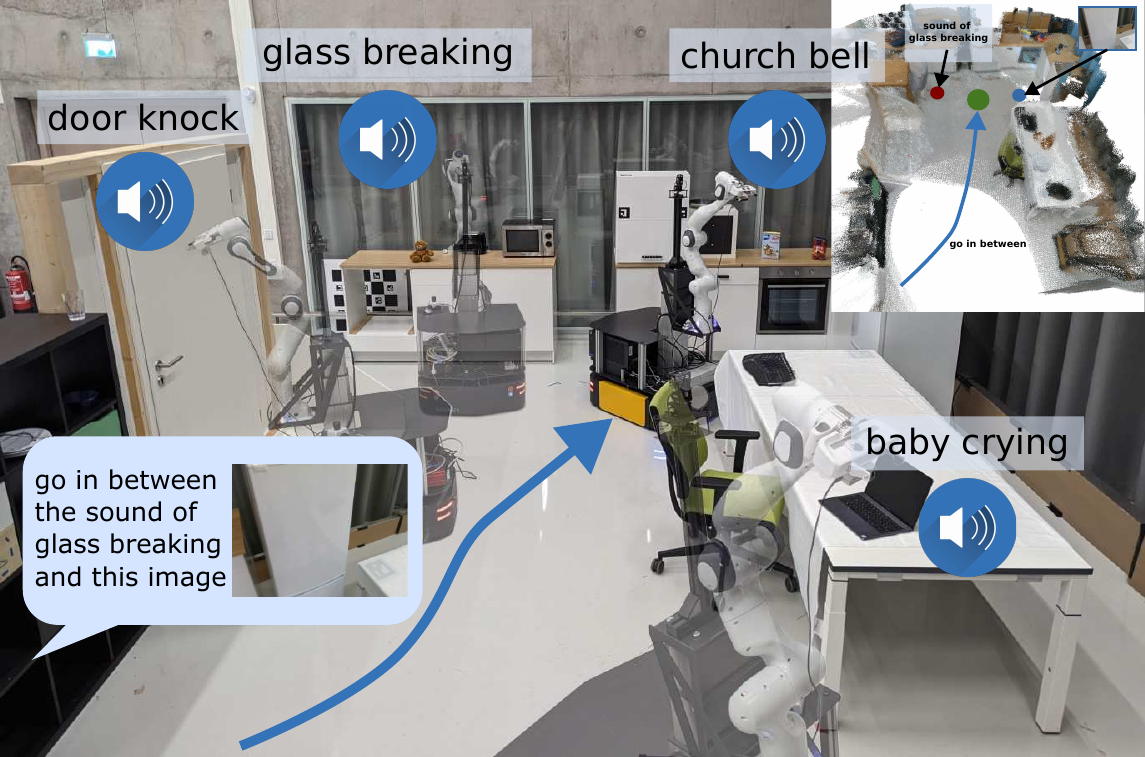}
	\caption{\small \textbf{AVLMaps} provide an open-vocabulary 3D map representation for storing cross-modal information
from audio, visual, and language cues. When combined with large language models, AVLMaps 
consumes multimodal prompts from audio, vision and language to solve zero-shot spatial goal navigation by effectively leveraging complementary information sources to disambiguate goals.}
	\label{fig:cover_lady}
	\vspace{-1em}
\end{figure}

To this end, we propose Audio-Visual-Language Maps, AVLMaps, a unified 3D spatial map representation for storing cross-sensing information from audio, visual, and language modalities. AVLMaps can be built from image and audio observations captured during reconstruction, by computing dense pre-trained features from open-vocabulary multimodal foundation models trained on Internet-scale data~\cite{li2022languagedriven,ghiasi2021open,guzhov2022audioclip} and fusing them into a shared 3D voxel grid representation. This map can then be used to index the locations of landmarks (or areas and regions of interest) in the environment via open-vocabulary multimodal queries, \eg abstract textual descriptions, images, or audio snippets -- enabling downstream applications including language-based goal-driven navigation, without domain-specific model finetuning. A key aspect of AVLMaps is that they extend prior multimodal mapping representations~\cite{huang23vlmaps, chen2022open, shafiullah2022clipfields} to include audio information, which allows robots to more often correctly disambiguate goal locations using sound -- \eg ``go to the table where you heard coughing'' in environments where there are multiple tables, etc. 
Additionally, when paired with large language models
(LLMs) we show that AVLMaps enable zero-shot multimodal \emph{spatial} goal localization, \eg ``Go in between the \{image of a refrigerator\} and the sound of breaking glass'' as in Fig.~\ref{fig:cover_lady}.

Extensive experiments in both simulated and real settings show that our system is capable of navigating to goal locations specified by, \eg natural language descriptions of sounds or visual landmarks -- and notably, can disambiguate multiple possible goal locations using multimodal information, (using object semantics to pinpoint one of the multiple possible sound goals, or using vision to pinpoint one of the multiple possible locations where similar objects were found) quantitatively better than baseline alternatives by up to 50\% in top-1 recall. Additional ablations further suggest these capabilities continue to naturally improve with better performing pre-trained audio-language foundation models such as AudioCLIP~\cite{guzhov2022audioclip}. AVLMaps are simple and effective in leveraging multiple multimodal foundation models together in tandem to reach broader language-driven robot navigation capabilities, but are also not without limitations -- we discuss these and avenues for future work. 

 \begin{figure*}[t]
	\centering
	\includegraphics[width=0.95\textwidth]{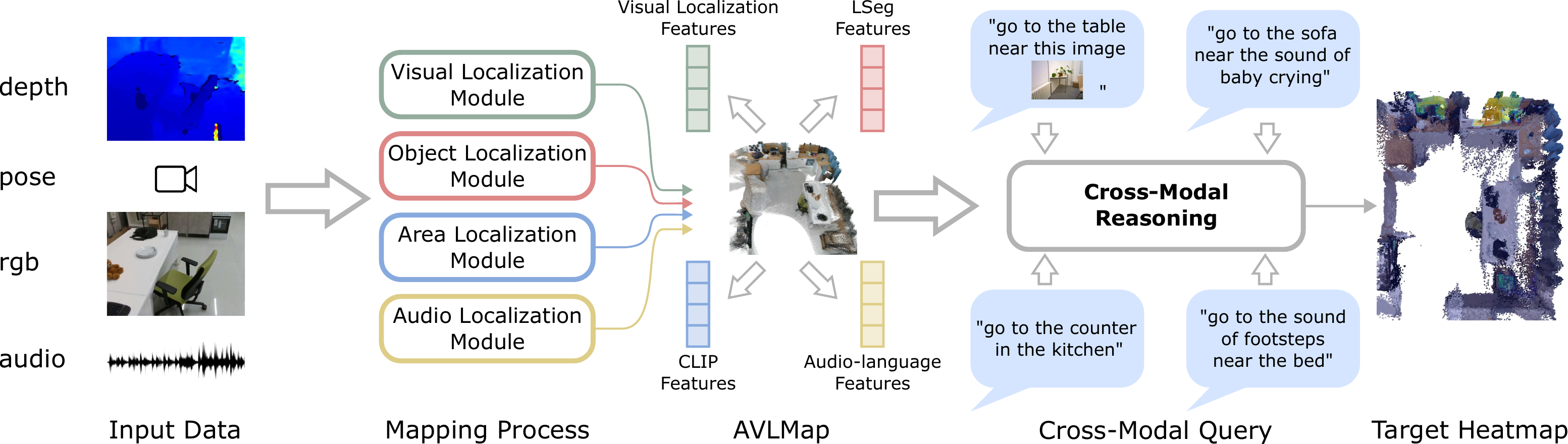}
	\caption{\small System overview. AVLMaps are constructed from RGB-D, audio, and odometry inputs, converting raw data into visual localization features, visual-language features, and audio-language features. During inference time, each module's output is unified with cross-modal reasoning, allowing users to query spatial location with multimodal information.}
	\label{fig:system_overview}
\end{figure*}

\section{Related Work}
\textbf{Semantic Mapping.}
 In recent years, the synergy of the traditional SLAM techniques and the advancements in vision-based semantic understanding  has led to augmenting 3D maps with semantic information~\cite{salas2013slam++, mccormac2017semanticfusion}. Stemming from the intuition of augmenting 3D points in the map with 2D segmentation results, previous works focus on either abstracting the map at object-level with a pose graph~\cite{mccormac2018fusion++} or an octree~\cite{xu2019mid} or modeling the dynamics of objects in the map~\cite{runz2018maskfusion}. Despite lifting the 3D reconstruction to a semantic level, these methods are restricted to a predefined set of semantic classes. Recent works like VLMaps~\cite{huang23vlmaps}, NLMap-SayCan~\cite{chen2022open}, OpenScene~\cite{peng2022openscene}, or CLIP-Fields~\cite{shafiullah2022clipfields} have shown that integrating visual-language features, generated by either pre-trained or fine-tuned models, into an occupancy map enables open-vocabulary object indexing with natural language, freeing the maps from fixed-size semantic categories. However, these works focus on visual perception to map and move through an environment, overlooking complementary sources of information such as acoustic signals. In contrast, AVLMaps integrate audio, visual, and language cues into a 3D map, equipping the agent with the ability to navigate to multiple types of multimodal goals and effectively disambiguate goals.

\textbf{Multimodal Navigation.}
Recent advances in simulation applications~\cite{habitat19iccv,kolve2017ai2,chen2020soundspaces,gan2021threedworld} have boosted research on multimodal navigation in two distinct directions: (i) vision-and-language navigation (VLN)~\cite{anderson2018vision,krantz2020beyond} where an agent needs to follow a natural language instruction towards the goal with visual input, and (ii) audio-visual navigation (AVN)~\cite{chen2020soundspaces} in which an agent should navigate to the sound source based on information from a binaural sensor and vision. Despite different degrees of success in both directions~\cite{fried2018speaker,guhur2021airbert,chen2020learning,younes2023catch,gan2020look,gan2022finding}, less attention has been paid to solving the navigation problem involving vision, language, and audio at the same time. The most relevant concept to our knowledge is from AVLEN~\cite{paul2022avlen}, which extends the AVN with a further query step, introducing a language instruction that helps with navigating to the sound source. In addition, most of the existing methods on AVN focus on approaching the sound without understanding its semantics. In our work, we propose a method to integrate both visual and sound semantics into the same map, enabling a robot to navigate to multimodal goals specified with either goal image or natural language like ``go to the sound of baby crying'', ``go to the table'' or multimodal prompts such as ``go to the \{image of a table\} where the sound of the microwave was heard''.

\textbf{Pre-trained Zero-shot Models in Robotics.}
Recent trends have shown that pre-trained models~\cite{radford2021learning,brown2020language} serve as powerful tools for robotic tasks including object detection and segmentation~\cite{kamath2021mdetr,gu2021open,li2021language}, robot manipulation~\cite{shridhar2022cliport,liang2022code,mees2022calvin,mees2022hulc,mees22hulc2,rosete2022tacorl}, and navigation~\cite{shah2022lm,gadre2022clip,chen2022nlmapsaycan,huang23vlmaps}.
Most related to our work are approaches like VLMaps~\cite{huang23vlmaps}, NLMap~\cite{chen2022nlmapsaycan}, and ConceptFusion~\cite{jatavallabhula2023conceptfusion}, all of which combine pre-trained visual-language models with a 3D reconstruction of the scene, enabling landmark indexing with natural language and downstream language-based planning tasks. However, little effort has been made to exploit the audio-language or visual-audio-language pre-trained models~\cite{elizalde2022clap,guzhov2022audioclip} in robotic tasks. Concurrent work ConceptFusion~\cite{jatavallabhula2023conceptfusion} demonstrates that audio can be used as queries to index locations in a visual-language map. Experiments in Sec.~\ref{sec:exp_cross_modal_goal_indexing} show that our method is complementary to ConceptFusion in that AVLMaps can excel in instances where ConceptFusion tends to struggle (indexing specific objects such as ``cabinet''), but likewise, ConceptFusion can compensate for cases where AVLMaps struggle (\eg localizing areas such as ``living room''). We show that the best-performing system uses both methods together in tandem.

\section{Method}
Our goal is to build an audio-visual-language map representation, in which object landmarks (``sofa''), areas (``kitchen''), audio semantics (``the sound of a baby crying''), or visual goals can be directly localized using natural language or target image. We propose AVLMaps as one such representation, which can be constructed by combining standard 3D reconstruction libraries with pre-trained visual-language models and audio-language models. We also propose a cross-modal reasoning method to disambiguate locations referring to targets from different modalities (``the sound of brushing teeth near the sink'', or ``the table near this image: \{image\}''). In the following subsections, we describe (i) how to build an AVLMap and use it for localizing different targets, (ii) how to disambiguate goal locations with multimodal information, (iii) how AVLMaps can be used with large language models (LLMs) for multimodal goal navigation, without additional data collection or model fine-tuning. The system pipeline is shown in Fig.~\ref{fig:system_overview}.

\subsection{Building an Audio Visual Language Map}
\label{subsec_building_avlmaps}
The key idea behind AVLMaps is to combine visual localization features, pre-trained visual-language features, and audio-language features with a 3D reconstruction. Given an RGB-D video stream with an audio track and odometry information, we utilize four modules to build a multimodal features database.

\textbf{Visual Localization Module.} The main purpose of this module is to localize a query image in our map. To achieve this goal, we follow a hierarchical localization scheme~\cite{sarlin2019coarse,sarlin2020superglue}. We first compute the NetVLAD~\cite{arandjelovic2016netvlad} global descriptors and SuperPoint~\cite{detone2018superpoint} local descriptors for all images from the RGB stream and store them with the corresponding depth and odometry. During inference, we compute the global and local descriptors for the query image in the same manner. By searching the nearest neighbor of the query NetVLAD features in the reference database, we can find a reference image as our candidate. Next, we use SuperGLUE~\cite{sarlin2020superglue} to establish key point correspondences between two images with their local features. With registered depth, we back-project reference key points into the 3D space and obtain the 3D-2D correspondences for the query key points. In the end, we can apply the Perspective-n-Point method~\cite{fischler1981random} to estimate the query camera pose relative to the reference camera, and thus obtain the global camera pose with the odometry of the reference camera. 

\textbf{Object Localization Module.} In this module, we follow the scheme in the work VLMaps~\cite{huang23vlmaps} to build a visual language map. The key idea is to exploit an open-vocabulary segmentation method (\eg LSeg~\cite{li2022languagedriven} or OpenSeg~\cite{ghiasi2021open}) for pixel-level feature generation from the RGB image and to associate these features with the back-projected depth pixels in the 3D reconstruction. During inference, we define a list of categories in natural language and encode them with the language encoder. We compute the cosine similarity scores between all point-wise features and language features and use an $\mathrm{argmax}$ operator to select the top-scoring points for a certain category in the map. Depending on the applications, the top-scoring 3D points for a certain category can be used as the target point cloud for manipulation tasks or can be projected onto a top-down map for navigation purposes.

\textbf{Area Localization Module.} While the object localization module is good at extracting object segments on the map, it falls short of localizing coarser goals such as regions (\eg ``the area of the kitchen''). This is because the visual encoder for generating pixel-aligned features is obtained by fine-tuning a pre-trained model on a segmentation dataset, leading to the notorious catastrophic forgetting effect. Therefore, the visual encoder is better at segmenting common objects while worse at recognizing general visual concepts~\cite{jatavallabhula2023conceptfusion}. To take advantage of both pre-trained and fine-tuned methods, we propose to build an extra sparse topological CLIP features map similar to~\cite{shah2022lm}. The idea is to compute the CLIP visual features~\cite{radford2021learning} for all images from the RGB stream and associate the features with corresponding poses. During inference, given the language concept like ``the area of a bedroom'', we compute the language features with the CLIP language encoder and the image-to-language cosine similarity scores. These similarity scores indicate how likely these images match the language description. The odometry together with the score of each image indicates the predicted location with a confidence value.

\textbf{Audio Localization Module.} In this module, we utilize the audio information from the input stream. The key idea is to compute the audio-lingual features with audio-language pre-trained models such as wav2clip~\cite{wu2022wav2clip}, AudioCLIP~\cite{guzhov2022audioclip} or CLAP~\cite{elizalde2022clap}. We first segment the whole audio clip into several segments with silence detection. Whenever the volume is above a threshold, we mark this time step as the starting point of a segment. Whenever the volume of the sound is not larger than this threshold for a certain duration, we end the segment. In the next step, we compute the audio features for each segment with pre-trained audio-language models and associate the features with the odometry at the specific segment. During inference, given a language description of the sound, like ``the sound of door knocks'', we encode the language into language features and compute the matching scores between the language and all audio segments in the same way as in the object localization module. The odometry associated with the top-scoring segment is the predicted location.

 \begin{figure}[t]
	\centering
	\includegraphics[width=1\columnwidth]{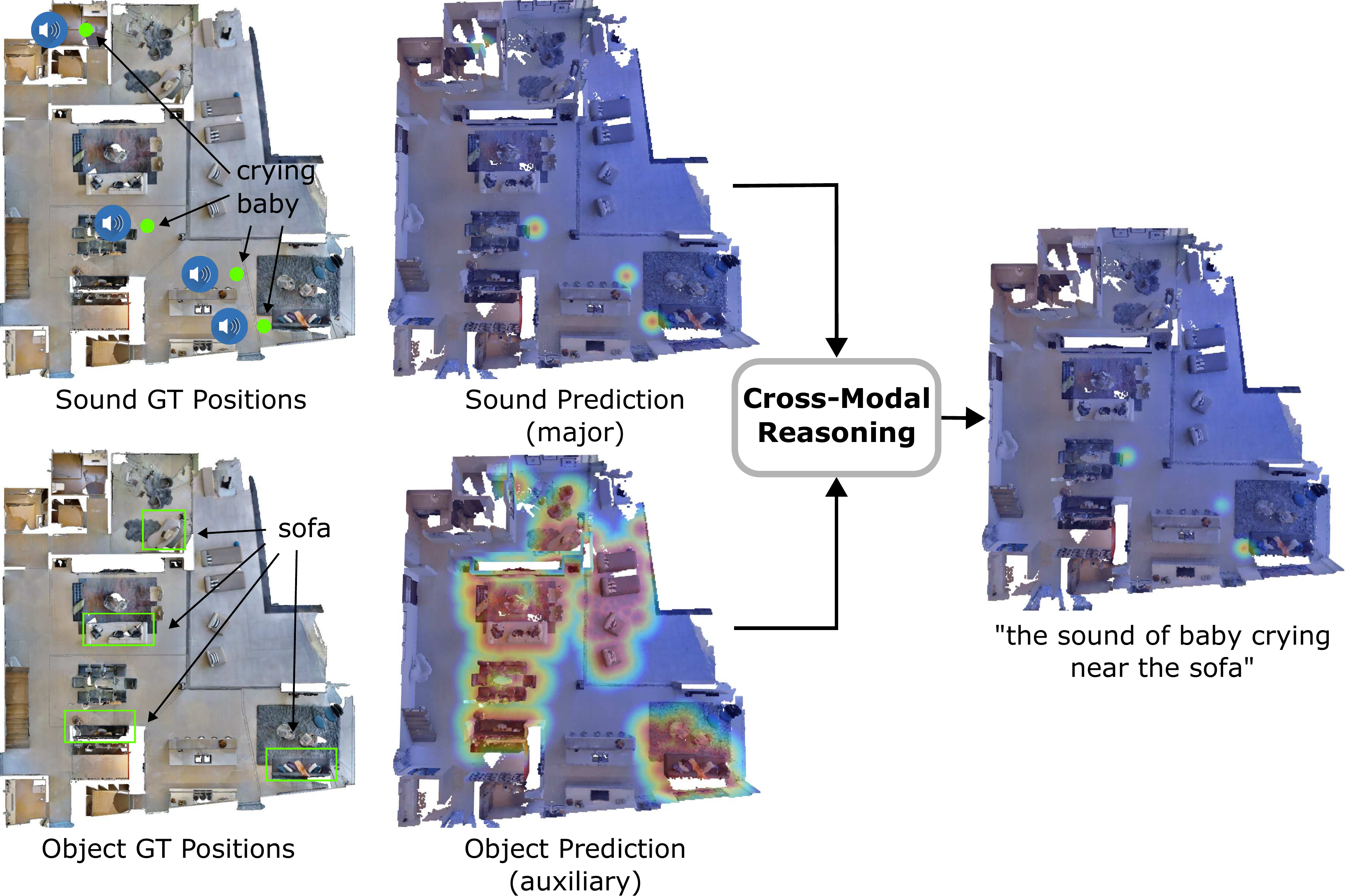}
	\caption{\small The key idea of cross-modal reasoning is converting the prediction from different modalities into heatmaps, and then fusing them with element-wise multiplication, effectively using complementary multimodal information to resolve ambiguous prompts.}
	\label{fig:cross_modal_fusion}
	\vspace{-1em}
\end{figure}

\subsection{Cross-Modality Reasoning}
\label{subsec_cross_modal_reasoning}
A key advantage of our method is its capability to disambiguate goals with additional information, even from different modalities. Given a specific query, each module introduced in the last section returns predicted spatial locations on the map in the form of 3D voxel heatmaps. A heatmap can be denoted as $\mathcal{H} \in [0, 1]^{\bar{H} \times \bar{W} \times \bar{Z}}$, where $\bar{H}$, $\bar{W}$ and $\bar{Z}$ represent the size of the voxel map and the value in each element represents the probability of being the target position. $\mathbf{p} = (x, y, z)^T, \{x,y,z \in \mathbb{Z}|1 \le x \le \bar{H}, 1 \le y \le \bar{W}, 1 \le z \le \bar{Z}\}$ is a voxel position in the map $\mathcal{H}$.

\textbf{Visual Localization Heatmap.} In the visual localization module, the predicted global camera location is denoted as $\mathbf{p}_v = (x_v, y_v, z_v)^T$. In the heatmap $\mathcal{H}_v$, we define the probability at $\mathbf{p}_v$ as 1.0, and the probability linearly decays around this location according to the distance on the top-down map:
\begin{equation}
\label{eq:vis_loc_heatmap}
\mathcal{H}_{v}(\mathbf{p}) = \mathrm{max} (1.0 - \epsilon \cdot dist_{xy}(\mathbf{p}, \mathbf{p}_v), 0)
\end{equation}
\begin{equation}
\label{eq:dist_xy}
dist_{xy}(\mathbf{p}, \mathbf{q}) = \sqrt{(p_x - q_x)^2 + (p_y - q_y)^2}
\end{equation}
where $\epsilon$ is the decay rate, and $dist_{xy}(\mathbf{p}, \mathbf{q})$ denotes the distance between 3D vectors $\mathbf{p}$ and $\mathbf{q}$ on the $xy$-plane.

\textbf{Object Localization Heatmap.} The object localization results are a list of points, denoted as $\{\mathbf{p}_{oi} = (x_{oi}, y_{oi}, z_{oi}) |  i = 1, 2, ..., N\}$ where $N$ is the total number of points for the target object. We define the probabilities for all these locations as 1.0 in heatmap $\mathcal{H}_o$, and the probability linearly decays around these locations based on the Euclidean distance:
\begin{equation}
\label{eq:min_d}
d_{min} (\mathbf{p}) = \mathrm{min} \{dist(\mathbf{p}, \mathbf{p}_{oi}) |  i = 1, 2, ..., N\}
\end{equation}
\begin{equation}
\label{eq:obj_loc_heatmap}
\mathcal{H}_{o}(\mathbf{p}) = \mathrm{max} (1.0 - \epsilon \cdot d_{min} (\mathbf{p}), 0)
\end{equation}
where $d_{min} (\mathbf{p})$ denotes the minimal distance between $\mathbf{p}$ and all object points $\{\mathbf{p}_{oi} |  i = 1, 2, ..., N \}$, $dist (\mathbf{p}, \mathbf{q})$ denotes the Euclidean distance between $\mathbf{p}$ and $\mathbf{q}$.

\textbf{Area Localization Heatmap.} The area localization results are a list of position-confidence pairs, denoted as $\{(\mathbf{p}_{ai}, s_{ai}) | i = 1, 2, ..., M\}$ where $M$ is the total number of frames in the input RGB-D stream. The scores $s_{ai}$ are normalized between 0 and 1. We define the probability for each point $\mathbf{p}_{ai}$ on the heatmap $\mathcal{H}_a$ as its score $s_{ai}$, and the probability linearly decays around the point on the $xy$-plane direction:
\begin{equation}
\label{eq:area_loc_heatmap}
\mathcal{H}_{a}(\mathbf{p}) = \mathrm{max} (\mathrm{max} \{s_{ai} - \epsilon \cdot dist_{xy}(\mathbf{p}, \mathbf{p}_{ai}) |  i = 1, 2, ..., M\}, 0)
\end{equation}
where the $\mathrm{max}$ operator for the curly brackets means taking the highest probability when a location is inside the affected regions for several $\mathbf{p}_{ai}$.

\textbf{Audio Localization Heatmap.} The audio localization results are similar to those of the area localization module. The position-score pairs are denoted as $\{(\mathbf{p}_{si}, s_{si}) | i = 1, 2, ..., K\}$ where $K$ is the total number of sound segments in the input video stream. The heatmap $\mathcal{H}_s$ is defined as:
\begin{equation}
\label{eq:sound_loc_heatmap}
\mathcal{H}_{s}(\mathbf{p}) = \mathrm{max} (\mathrm{max} \{s_{si} - \epsilon \cdot dist_{xy}(\mathbf{p}, \mathbf{p}_{si}) |  i = 1, 2, ..., K\}, 0)
\end{equation}

\textbf{Cross-Modal Reasoning.} The main idea of cross-modal reasoning is shown in Fig.~\ref{fig:cross_modal_fusion}. We treat the predictions from four modules as four modalities. When there are several queries referring to different modalities, we compute the respective heatmaps first and then perform element-wise multiplication among all heatmaps:
\begin{equation}
\label{eq:heatmap_mul}
\mathcal{H}_{target} = \mathcal{H}_{1} \odot \mathcal{H}_{2} \odot ... \odot \mathcal{H}_{L} 
\end{equation}
where $\odot$ is the element-wise multiplication operator, and $L$ is the total number of referred modalities. We extract the position on the target heatmap $\mathcal{H}_{target}$ that has the highest probability as the predicted location.

When we compute the heatmaps, there is always a primary heatmap while others are auxiliary ones. For example, in the query ``the chair near the sound of crying'', the heatmap for ``the chair'' is the primary heatmap, while the heatmap for ``the sound of crying'' is the auxiliary. We set the decay rate for the primary heatmap higher (\eg 0.1 in this work) since we want to know the exact location of the target, while tuning the decay rate for the auxiliary heatmap lower (\eg 0.01) as having a broader affect area to narrow down major targets is desirable.

\subsection{Multimodal Goal Navigation from Language}\label{subsec_multimodal_nav_from_language}
In the setting of multimodal goal navigation from language, the agent is given language descriptions of targets from different modalities (\eg sound, image, and object) and is required to plan paths to them. While most of the previous navigation methods focus mainly on a specific type of goal, we unify these tasks with the help of large language models (LLMs). Specifically, we use an LLM to interpret the natural language commands and synthesize API calls combined with simple logic structures in the form of executable python code~\cite{liang2022code,huang23vlmaps,mees22hulc2}. For heatmap generation, we implement interfaces \textit{\textbf{get\_major\_map}(obj=None, sound=None, img=None)} and \textit{\textbf{get\_map}(obj=None, sound=None, img=None)}. They take object name, sound name, or image as input and output heatmaps indicating the locations of targets. The \textit{\textbf{get\_major\_map}} generates heatmaps with higher decay rate while \textit{\textbf{get\_map}} with lower decay rate. To support image prompt, we add image path in the language query like ``the image /path/to/image.png'' and use LLMs to call the image loading API. Some examples of prompts and queries are shown below (prompt in {\color{prompt-gray}gray}, input task commands in \command{green}, and generated outputs are \colorbox{highlight}{highlighted}):
\lmp{
\prompt{
\# move to the middle of the sound of cat meowing and\\
the image: /path/to/image.png\\
img = robot.load\_image("/path/to/image.png")\\
sound\_map = robot.get\_major\_map(sound="cat meowing")\\
img\_map = robot.get\_major\_map(img=img)\\
pos1 = robot.get\_max\_pos\_3d(sound\_map)\\
pos2 = robot.get\_max\_pos\_3d(img\_map)\\
pos = (pos1 + pos2) / 2\\
robot.move\_to(pos)\\
\# move to the window next to the sound of\\
glass breaking\\
obj\_map = robot.get\_major\_map(obj="window")\\
sound\_map = robot.get\_map(sound="glass breaking")\\
fuse\_map = obj\_map * sound\_map\\
pos = robot.get\_max\_pos\_3d(fuse\_map)\\
robot.move\_to(pos)\\
...
}}
\speciallmp{
\command{\# move in between the image ./006899.png and the}\\
\command{backpack near the sound of glass breaking}\\
\hlcode{img = robot.load\_image("./006899.png")}\\
\hlcode{img\_map = robot.get\_major\_map(img=img)}\\
\hlcode{obj\_map = robot.get\_major\_map(obj="backpack")}\\
\hlcode{sound\_map = robot.get\_map(sound="glass breaking")}\\
\hlcode{fuse\_map = obj\_map * sound\_map}\\
\hlcode{pos1 = robot.get\_max\_pos\_3d(img\_map)}\\
\hlcode{pos2 = robot.get\_max\_pos\_3d(fuse\_map)}\\
\hlcode{pos = (pos1 + pos2) / 2}\\
\hlcode{robot.move\_to(pos)}
}
\section{Experiments}
\label{sec:experiments}

\subsection{Simulation Setup}\label{sec:exp_sim_setup}
\textbf{Exerimental setup.} We use the Habitat simulator~\cite{habitat19iccv,szot2021habitat} with the Matterport3D dataset~\cite{Matterport3D} for the evaluation of multimodal navigation tasks. The dataset contains a large set of apartment-level realistic indoor scenes that help evaluate the generalization capabilities of navigating agents. For mapping purposes, we manually collect RGB-D video streams in the simulator across 10 different scenes and add random audio tracks to the videos to simulate the audio sensing modality. All audio comes from the validation fold (Fold-1) of the ESC-50 dataset~\cite{piczak2015esc}, which contains 50 categories of common sounds. In navigation tasks, the robot has four actions to take: \textbf{move forward 0.1 meters}, \textbf{turn left 5 degrees}, \textbf{turn right 5 degrees}, and \textbf{stop}. In sequential goal setting, the robot is required to navigate to a sequence of goals and take the \textbf{stop} action when it reaches each subgoal. When the stop position is less than 1 meter from the ground truth position, the subgoal is considered successfully finished.

\textbf{Tasks collection.} In multimodal goal navigation tasks in Sec.~\ref{sec:exp_multi_modal_goal_nav}, we consider three kinds of goals: image goals, object goals, and sound goals. For image goals, we randomly sample positions and orientations on the top-down map and render images as targets. For object goals, we access the metadata (\eg bounding boxes and semantics) from the Matterport3D dataset and sample a list of categories in each scene as queries. For sound goals, we randomly sample sound classes of audio merged with the mapping videos as targets, treating the video frame positions as the ground truth. 

In cross-modal goal indexing tasks in Sec.~\ref{sec:exp_cross_modal_goal_indexing}, we collect three types of datasets: 
\begin{itemize}[leftmargin=*]
  \item \textbf{Visual-Object cross-modal indexing} We manually select image-object pairs on the top-down map for localizing ``an object X near the image Y''.
  \item \textbf{Area-Object cross-modal indexing} We access the region and object metadata (\eg bounding boxes and semantics) from the Matterport3D dataset to automatically generate a list of object-region pairs. This dataset is for localizing ``an object X in the area of Y''.
  \item \textbf{Object-Sound cross-modal indexing} We manually insert several sounds of the same kind into a scene and select for each sound location a nearby object for disambiguation. The query is ``a sound X near the object Y''.
\end{itemize}

In cross-modal goal navigation in Sec.~\ref{sec:exp_cross_modal_nav}, we randomly sample starting pose in 10 scenes and treat the visual-object and object-sound cross-modal goals in Sec.~\ref{sec:exp_cross_modal_goal_indexing} as navigation goals.

\subsection{Multimodal Goal Navigation}\label{sec:exp_multi_modal_goal_nav}
\textbf{Sound goal navigation.} We first test AVLMaps in sound goal navigation tasks. We collect 200 sequences of sound goals in 10 different scenes. In each sequence, there are 4 sound categories that require the robot to reach. The results are shown in Tab.~\ref{table:multi_audio_nav}. We generate AudioCLIP~\cite{guzhov2022audioclip} features with our audio localization module and match all audio with the target sound category in the embedding space, similar to a text-to-audio retrieval setup. Then the agent plans a path to the audio position. We tested different ranges of sound categories inserted into the map. The full list of sound categories in each major class can be found in the link\footnote{\href{https://github.com/karolpiczak/ESC-50}{https://github.com/karolpiczak/ESC-50}}. The results show that our agent manages to recognize sound goals and navigate with a 77.5\% success rate.
\begin{table}[h]
  \setlength\tabcolsep{5.2pt}
  \centering
  \begin{tabular}{lccccc}
  \toprule
\multirow{2}[1]{*}{Tasks}  & \multicolumn{4}{l}{No. Subgoals in a Row} & Independent      \\                
                    \cmidrule(lr){2-5}
        &   1             & 2           & 3           & 4    &  Subgoals\\
\midrule
Domestic Sound & 59.5         & 33.0      & 15.5     & 7.0     & 62.5     \\
+ Human Sound & 69.5         & 47.0      & 36.5     & 23.0     & 72.38     \\
+ Animal Sound & 74.5         & 58.5      & 45.5     & 33.0     & 77.5     \\

  \bottomrule
  \end{tabular}
  \caption{The success rate (\%) of sound goal navigation with AVLMaps.}
  \label{table:multi_audio_nav}
  \vspace{-1em}
\end{table}

\textbf{Visual and object goals navigation.} We then test AVLMaps with visual and object goal navigation tasks. The agent is given an image and two object categories in the language in one sequence of tasks and asked to navigate to the image goal and two object goals in sequence. In 200 sequences of tasks in 10 scenes, the success rate is reported in Tab.~\ref{table:visual_object_goal_nav}. The results show that our method enables the agent to navigate to goals from different modalities.
\begin{table}[h]
  \setlength\tabcolsep{5.2pt}
  \centering
  \begin{tabular}{lccccc}
  \toprule
\multirow{2}[1]{*}{Tasks}  & \multicolumn{3}{l}{No. Subgoals in a Row} & Independent      \\                
                    \cmidrule(lr){2-4}
               &   1             & 2           & 3              &  Subgoals\\
\midrule
AVLMaps (Ours) & 71.5            & 40.5        & 25.0           & 47.4     \\

  \bottomrule
  \end{tabular}
  \caption{The success rate (\%) of multimodal goal navigation with AVLMaps. The agent is required to navigate to one visual goal, and two object goals in sequence.}
  \label{table:visual_object_goal_nav}
  \vspace{-1.4em}
\end{table}

\subsection{Cross-Modal Goal Indexing}\label{sec:exp_cross_modal_goal_indexing}
When we refer to a goal with language, it is likely that the goal can be found in more than one place in the environment. A major strength of our method is that it can disambiguate goals with multimodal information. In this experiment, we will show the cross-modal goal reasoning capability of AVLMaps.

\textbf{Area-Object goal indexing.} In this setup, we use an area description to disambiguate the object goal. We collected 100 indexing tasks in 10 scenes. Each task consists of an object category and a region category (\eg ``living room'', ``kitchen'', ``dining room'', ``bathroom'' etc.). The agent needs to predict the correct object location which is inside the region. The top-1 recall with different distance tolerance is reported in Tab.~\ref{table:area_object_indexing_results}. We can notice that VLMaps~\cite{huang23vlmaps} struggles to find the goal in the correct region because VLMaps integrates visual-language features from the encoder fine-tuned on the instance segmentation dataset, improving its segmentation performance on common objects while dropping its ability to recognize more general concepts like regions. In contrast, ConceptFusion integrates pre-trained CLIP features into the map without fine-tuning, enabling it to recognize general concepts including regions, and thus the indexing results are improved.

\begin{table}[h]
  \setlength\tabcolsep{5.2pt}
  \centering
  \begin{tabularx}{\columnwidth}{lccccc}
  \toprule

\multirow{2}{*}{Method}                                            & \multicolumn{4}{c}{Recall@1 (\%)}                                 & \multirow{2}{*}{\begin{tabular}[c]{@{}c@{}}Average\\ min. distance \\ (m)\end{tabular}} \\ \cmidrule{2-5}
                                                                   & \textless{}0.5m & \textless{}1m & \textless{}1.5m & \textless{}2m & \\
\midrule

\begin{tabular}[c]{@{}c@{}}baseline \\ (VLMaps)\end{tabular}        & 5.56            & 7.78          & 13.33           & 17.78         & 8.22 \\ 
\begin{tabular}[c]{@{}c@{}} + ConceptFusion\end{tabular}   & 12.22           & 13.33         & 16.67           & 21.11         & 7.60 \\ 
\begin{tabular}[c]{@{}c@{}} + CLIP sparse map \\ (Ours)\end{tabular} & \textbf{15.56}           & \textbf{24.44}         & \textbf{31.11}           & \textbf{35.56}         & \textbf{6.17} \\ 
\begin{tabular}[c]{@{}c@{}} + GT region map\end{tabular}   & 37.78           & 44.44         & 55.56           & 61.11         & 2.62 \\ 

  \bottomrule
  \end{tabularx}
\caption{The recall (\%) of area-object cross-modal indexing. }
  \label{table:area_object_indexing_results}
  \vspace{-1em}
\end{table}

\textbf{Object-Sound goal indexing.} In this setting, we use object goals to disambiguate sound goals. We collected 119 indexing tasks, each of which consist of a sound category and a nearby object category. Each sound category in a scene can be heard at more than 1 location, introducing ambiguity to the localization scenario. The recall is reported in Tab.~\ref{table:object_sound_indexing_results}. With the combination of object and audio localization modules, our method largely increases the recall rate for localizing the correct sound goal position in ambiguous scenarios. 
\begin{table}[h]
  \setlength\tabcolsep{5.2pt}
  \centering
  \begin{tabular}{lccccc}
  \toprule
\multirow{2}{*}{Method}                                             & \multicolumn{4}{c}{Recall@1 (\%)}                                 & \multirow{2}{*}{\begin{tabular}[c]{@{}c@{}}Average\\ min. distance (m)\end{tabular}} \\ \cmidrule{2-5}
                                                                    & \textless{}0.5m & \textless{}1m & \textless{}1.5m & \textless{}2m &                                                                                      \\ 
\midrule
\begin{tabular}[c]{@{}c@{}}baseline \\ (wav2clip) \end{tabular} & 8.40           & 10.08         & 10.92           & 14.29         & 8.52                                                                                 \\
\begin{tabular}[c]{@{}c@{}}baseline \\ (AudioCLIP) \end{tabular} & 26.05           & 35.29         & 36.97           & 42.01         & 5.04                                                                                 \\
\begin{tabular}[c]{@{}c@{}}VLMaps + \\ wav2clip\end{tabular}         & 24.37           & 30.25         & 33.61           & 38.66         & 6.27                                                                                 \\ 
\begin{tabular}[c]{@{}c@{}}VLMaps + \\ AudioCLIP \\ (Ours)\end{tabular} & \textbf{53.78}           & \textbf{65.55}         & \textbf{67.23 }          & \textbf{70.59}         & \textbf{2.74}                                                                                 \\ 
\bottomrule
\end{tabular}
\caption{The recall (\%) of object-sound cross-modal indexing. }
  \label{table:object_sound_indexing_results}
  \vspace{-1em}
\end{table}

\textbf{Visual-Object goal indexing.} In visual-object goal indexing tasks, visual clues are used to resolve ambiguity. Given an object category and an image, our method can localize the correct object near the image position with over 60\% of recall for 0.5 meters distance tolerance, as is shown in Tab.~\ref{table:visual_object_indexing_results}.

\begin{table}[h]
  \setlength\tabcolsep{5.2pt}
  \centering
  \begin{tabularx}{\columnwidth}{lccccc}
  \toprule
\multirow{2}{*}{Method}  & \multicolumn{4}{c}{Recall@1 (\%)} & \multirow{2}{*}{\begin{tabular}[c]{@{}c@{}}Average\\ min. distance \\ (m)\end{tabular}} \\ \cmidrule{2-5}& \textless{}0.5m & \textless{}1m & \textless{}1.5m & \textless{}2m & \\ 
\midrule
\begin{tabular}[c]{@{}c@{}}VLMaps w/o \\ visual localization\end{tabular}         & 7.55           & 9.43         & 11.32           & 11.94         & 11.22 \\ 
\begin{tabular}[c]{@{}c@{}}VLMaps w/ \\ visual localization \\ (Ours) \end{tabular} & \textbf{62.26}    &\textbf{66.67}         & \textbf{70.44}           & \textbf{72.32}         & \textbf{3.11} \\ 

\bottomrule
\end{tabularx}
\caption{The recall (\%) of visual-object cross-modal indexing.}
  \label{table:visual_object_indexing_results}
  \vspace{-1em}
\end{table}

\subsection{Multimodal Ambiguous Goal Navigation}\label{sec:exp_cross_modal_nav}
In this part, we test our method with ambiguous goal navigation tasks. we collect 119 sequences of tasks. In each task, the agent is required to navigate to an ambiguous sound goal (\ie ``move to the sound of baby crying near the sofa'') and an ambiguous object goal (\ie ``move to the counter near \{image of the kitchen\}'') sequentially. The category of the object near the sound and the image taken near the object are provided. These tasks require the agent to reason across different modalities to accurately localize the target. We consider two single-modality baselines: VLMaps~\cite{huang23vlmaps} and AudioCLIP~\cite{guzhov2022audioclip} and one multimodal baseline. The multimodal baseline uses VLMaps as the object localization module, wav2clip~\cite{wu2022wav2clip} as the audio localization module and the same visual localization module as our method. The results are shown in Tab.~\ref{table:cross_modal_nav}. We observe that AVLMaps navigates to cross-modal goals with 24.2\% higher success rate to ambiguous sound goals and with 2.1\% higher success rate to ambiguous object goals compared to the alternative multimodal baseline.

\begin{table}[h]
  \setlength\tabcolsep{5.2pt}
  \centering
  \begin{tabular}{lccccc}
  \toprule
\multirow{2}[1]{*}{Tasks}  & \multicolumn{2}{l}{No. Subgoals in a Row} & Sound  & Object    \\                
                    \cmidrule(lr){2-3}
               &   1             & 2                        &  Goals   & Goals \\
\midrule
VLMaps~\cite{huang23vlmaps}         &    -            &       -                  &    -      &   27.1       \\
AudioCLIP~\cite{guzhov2022audioclip}      &    -            &       -           &    16.9       &   -    \\
VLMaps + wav2clip & 22.0    & 12.7        & 22.0 & 53.4     \\
VLMaps + AudioCLIP (Ours) & \textbf{46.2}    & \textbf{28.6}        & \textbf{46.2} & \textbf{55.5}     \\

  \bottomrule
  \end{tabular}
  \caption{The success rate (\%) of multimodal ambiguous goal navigation with AVLMaps. The agent is required to navigate to one ambiguous sound goal and one ambiguous object goal sequentially.}
  \label{table:cross_modal_nav}
  \vspace{-1em}
\end{table}

\subsection{Real World Experiment}\label{sec:exp_real_world}

 \begin{figure}[t]
	\centering
	\includegraphics[width=1\columnwidth]{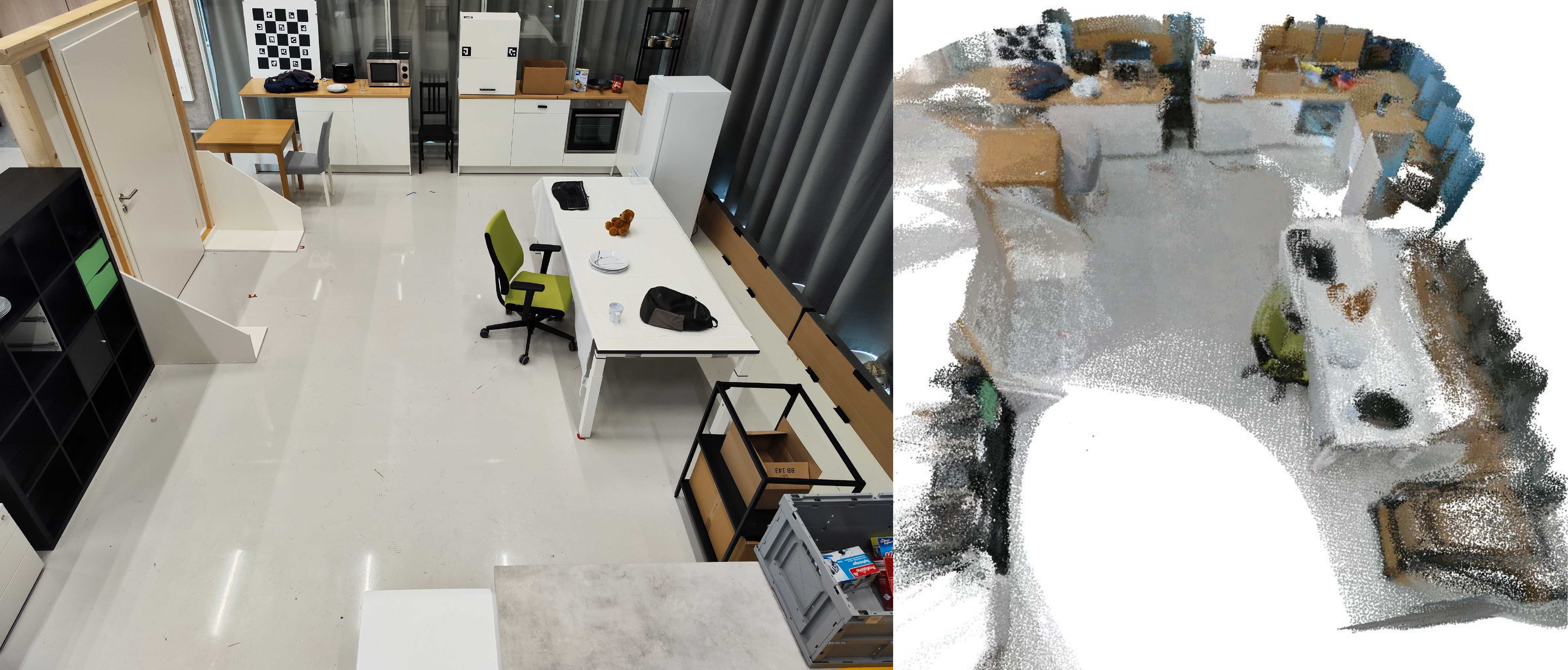}
	\caption{\small Real-world experiments are conducted in a room with multiple ambiguous goals such as tables, chairs, backpacks, and paper boxes (left). We leverage dense SLAM techniques to build a 3D reconstruction of the scene from RGB-D camera data into which we anchor features from multiple foundation models (right).}
	\label{fig:exp_imbit_2}
	\vspace{-1em}
\end{figure}

\textbf{Robot setup.} In the real-world experiment setting, we use a mobile robot equipped with a Ridgeback omnidirectional platform from Clearpath Robotics as the mobile base, and a Panda manipulator from Franka Emika. We mount a RealSense D435 RGB-D camera at the gripper of the Panda manipulator. During the mapping, we run a LiDAR localizer to provide the odometry for the robot base and derive the camera pose through the forward kinematics of the robot arm.

\begin{figure}[b]
	\centering
	\includegraphics[width=0.8\columnwidth]{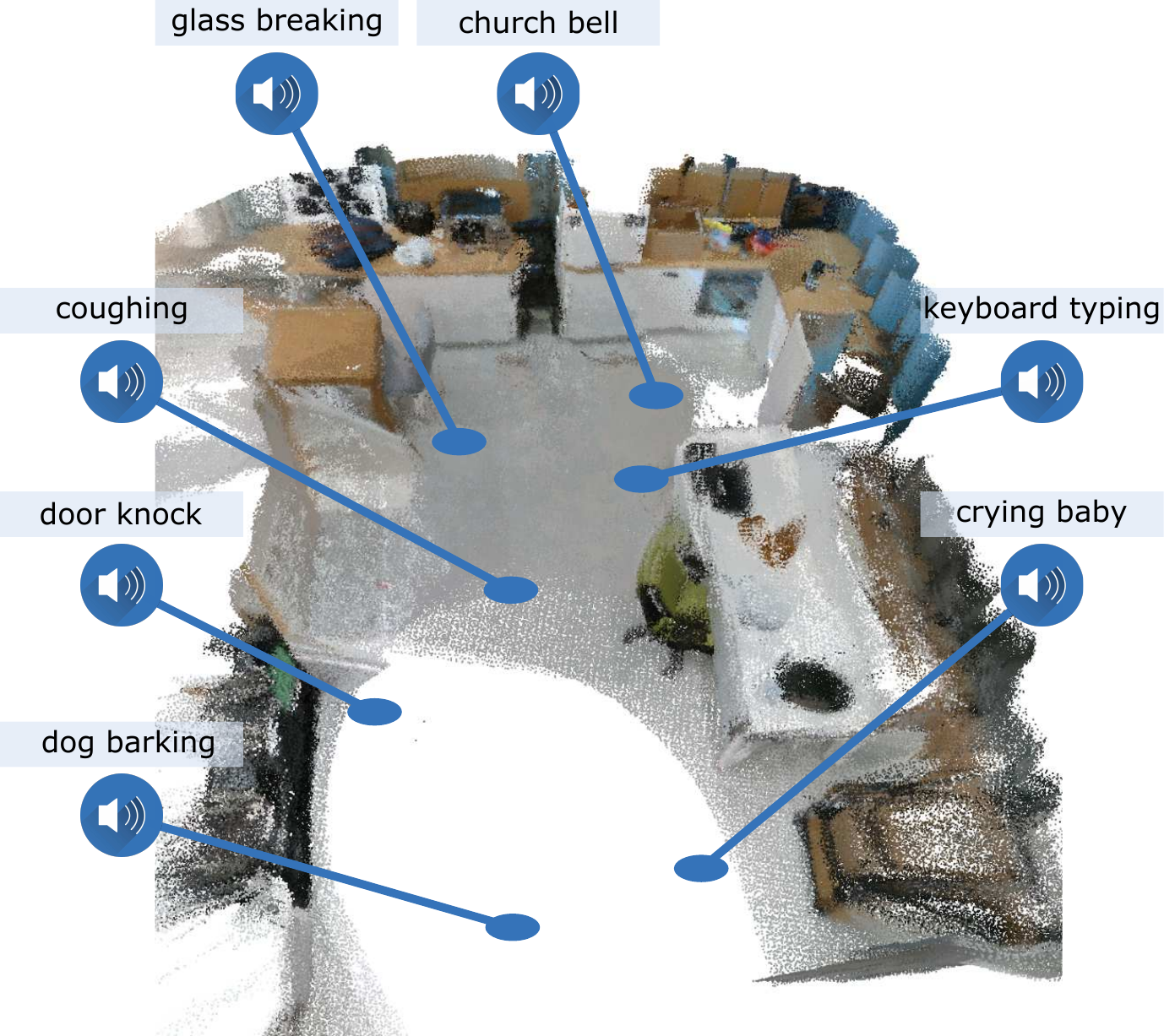}
	\caption{\small We artificially insert sounds with different semantics at locations shown in the image. Different sounds are played when the robot moves to these locations during mapping. Sounds are sampled from the ESC-50 dataset.}
	\label{fig:exp_real_world_sound_pos}
	\vspace{-1em}
\end{figure}

\textbf{Environment setup.} We choose a room with multiple ambiguous goals such as tables, chairs, paper boxes, counters, and backpacks which are shown in Fig.~\ref{fig:exp_imbit_2}. We control the robot in this environment and record RGB-D video. Then we artificially add sounds to the RGB-D video when the robot moves to certain locations. The sound locations are shown in Fig.~\ref{fig:exp_real_world_sound_pos}. After collecting the data, we run the AVLMaps mapping offline. For navigation tasks, we provide the AVLMaps and the language instruction as input. The robot parses the instruction (Sec.~\ref{subsec_multimodal_nav_from_language}) and executes the generated python code for goal indexing and planning. We use the ROS navigation package~\cite{quigley2009ros} for global and local planning. To avoid including noise proceeding from the own robots operation, we preprocess sound inputs with background noise substraction.

\begin{figure}[b]
	\centering
	\includegraphics[width=1\columnwidth]{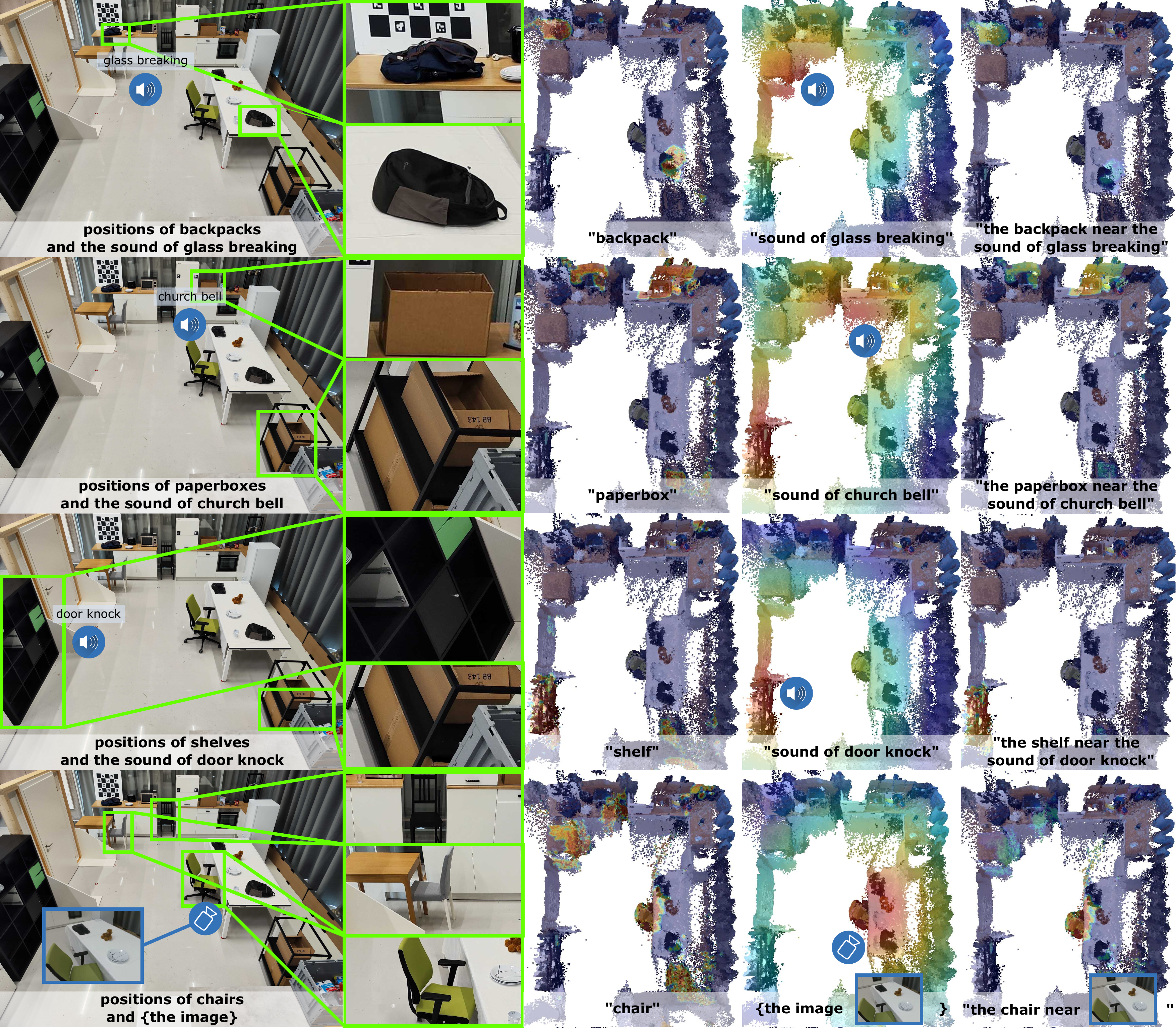}
	\caption{\small Visualization of example heatmaps in AVLMaps for multimodal goal reasoning for ambiguous object goals. The first column shows the positions of ambiguous objects (green bounding boxes) and the location of a sound (the icon of a speaker) or an image (the icon of a camera). The second column shows the zoom-in view of ambiguous objects in the scene. The third column shows the predicted 3D heatmap for the object. The fourth column shows the heatmap for the extra modality. The final column shows the fused heatmap after cross-modal reasoning. Sounds are artificially inserted into the scene for benchmarking and evaluation. The locations of sounds are not sound source locations but the places where the sounds were heard. The heatmap is shown in JET color scheme (red means the highest score and blue means the lowest score).}
	\label{fig:real_object_heatmap}
	\vspace{-1em}
\end{figure}

\textbf{Multimodal Spatial Goal Reasoning and Navigation with Natural Language.}  We design 20 language-based multimodal navigation tasks, asking the robot to navigate to sounds, images, and objects. We report an overall success rate of 50\%. We also design an evaluation consisting of 10 multimodal spatial goals. The agent needs to reason across object, sound, image and spatial concepts. An example is ``navigate in between the backpack near the sound of glass breaking and \{the image of a fridge\}''. In the end, 6 out of 10 tasks were successfully finished. We show in Fig.~\ref{fig:real_object_heatmap} the process of resolving ambiguities in the scene. There are different ambiguous objects in the scenes including paper boxes, backpacks, shelves, tables, chairs, and plates. The first and the second columns in Fig.~\ref{fig:real_object_heatmap} show the ground truth positions of the target objects and sounds. The third and fourth columns show that AVLMaps can accurately localize objects, sounds, and visual goals in the form of 3D heatmaps. The final column shows that our method can correctly narrow down targets in spite of object ambiguities. We can observe from the figure that AVLMaps can accurately localize ambiguous concept with language, audio and image. We observe that the failures come from the composition of the imperfection of different modules. For example, the object localization module (\eg VLMaps) fails to recognize rare objects like various toys. It also mistakes some shelves for chairs. Similar failures happen in audio localization module. In the second row and the fourth column in Fig.~\ref{fig:real_object_heatmap}, the church bell sound should be at the top-right corner but the module also gives high score for the sound heard at bottom-left.

\section{Conclusion and Limitations}
In this paper, we introduced AVLMaps, a unified 3D spatial map representation for storing cross-modal information from audio, visual, and language cues. When combined with large language models, AVLMaps 
consumes multimodal prompts from audio, vision and language to solve zero-shot spatial goal navigation by effectively leveraging complementary information sources to disambiguate goals. Experiments show that our cross-modal reasoning method can largely improve the target indexing accuracy when there are ambiguous goals compared to baselines.
While the AVLMaps approach is quite capable, it does have a number of limitations. AVLMaps are sensitive to the noise in the recorded audio. In addition, AVLMaps assume a static environment throughout their lifetime, including the mapping period and the application phase. When the layout of objects in the same room changes, the agent will still take action according to its memory. An interesting future direction is to integrate the life-long learning ability into the agent, automating multimodal spatial learning.






\section*{ACKNOWLEDGMENT}
This work has been supported partly by the German Federal Ministry of Education and Research under contract 01IS18040B-OML. We thank the BrainLinks-BrainTools center for providing the mobile manipulator used in our study for the real-world experiment.


\bibliographystyle{IEEEtran}
\bibliography{references}


\end{document}